\documentclass[journal]{IEEEtai}
\usepackage[colorlinks,urlcolor=blue,linkcolor=blue,citecolor=blue]{hyperref}
\usepackage{graphicx}

\usepackage{amsmath, amssymb, amsfonts}
\usepackage{url}
\usepackage{cite}
\usepackage{bm}
\usepackage{arydshln}
\usepackage{algorithmic}
\usepackage{algorithm}
\usepackage{authblk}
\usepackage{lscape}

\newcommand{\m}{\mathrm}
\newcommand{\p}{\prime}
\newcommand{\ov}{\overline}
\newcommand{\argmin}{\mathop{\rm argmin}\limits}
\newcommand{\argmax}{\mathop{\rm argmax}\limits}
\newcommand{\newmax}{\mathop{\rm max}\limits}

\begin{document}
\title{{\Large Multi-rules mining algorithm for combinatorially exploded decision trees with modified Aitchison-Aitken function-based Bayesian optimization}}

\author{Yuto Omae, Masaya Mori, and Yohei Kakimoto
\thanks{This work was supported in part by JSPS (Grant No. 23K11310).}
\thanks{Y. Omae, M. Mori, and Y. Kakimoto are with College of Industrial Technology, Nihon University, Chiba, 275-8575, Japan (e-mail: oomae.yuuto@nihon-u.ac.jp).}
}
\markboth{Preprint version (2023)}
{Y. Omae \MakeLowercase{\textit{et al.}}: MAABO-MT and GS-MRM algorithms}

\maketitle

\begin{abstract}
Decision trees offer the benefit of easy interpretation because they allow the classification of input data based on if--then rules.
However, as decision trees are constructed by an algorithm that achieves clear classification with minimum necessary rules, the trees possess the drawback of extracting only minimum rules, even when various latent rules exist in data.
Approaches that construct multiple trees using randomly selected feature subsets do exist. However, the number of trees that can be constructed remains at the same scale because the number of feature subsets is a combinatorial explosion.
Additionally, when multiple trees are constructed, numerous rules are generated, of which several are untrustworthy and/or highly similar.
Therefore, we propose ``MAABO-MT'' and ``GS-MRM'' algorithms that strategically construct trees with high estimation performance among all possible trees with small computational complexity and extract only reliable and non-similar rules, respectively.
Experiments are conducted using several open datasets to analyze the effectiveness of the proposed method.
The results confirm that MAABO-MT can discover reliable rules at a lower computational cost than other methods that rely on randomness.
Furthermore, the proposed method is confirmed to provide deeper insights than single decision trees commonly used in previous studies.
Therefore, MAABO-MT and GS-MRM can efficiently extract rules from combinatorially exploded decision trees.
\end{abstract}


\section{Introduction}
A decision tree is one of the supervised learning methods for classifying input data \cite{tree_quinlan1}. It offers the advantage of easy interpretation by analysts as data classification is based on a simple if--then rule, making it effective for acquiring knowledge from data \cite{tree_gini1} and promoting application to diverse domains \cite{tree_indust1, tree_agri2, tree_edu2}.
Decision trees are constructed by optimizing the Gini index \cite{tree_gini1, tree_gini_infog1} or information gain \cite{tree_quinlan1} and acquiring the minimum number of necessary rules for data classification, thereby proving consistency with the principle of Occam's Razor \cite{tree_occam1}.
However, only a small fraction of the large number of latent rules in data can be extracted.

A method to overcome the aforementioned problem includes multiple decision tree construction. For instance, multiple decision trees can be constructed using randomly selected features such as random forest (RF) \cite{rf_origin1}. Several studies have analyzed the internal structure of RF \cite{rf_ex1, rf_ex2, rf_ex3} and extracted multiple rules through its application. However, as the number of combinations of randomly selected features is enormous, the constructed decision trees can be inappropriate.
Additionally, decision trees with appropriate rules can be overlooked before construction. Therefore, the strategical construction of multiple decision trees with good performance using methods with low computational costs is necessary, and random selection should be avoided.
Additionally, decision trees can adopt meaningless noise features owing to parameter optimization (Appendix 1).
Therefore, extracting rules from decision trees using randomly selected features is perilous.

\begin{figure*}[t]
 \centering
 \includegraphics[scale=0.17]{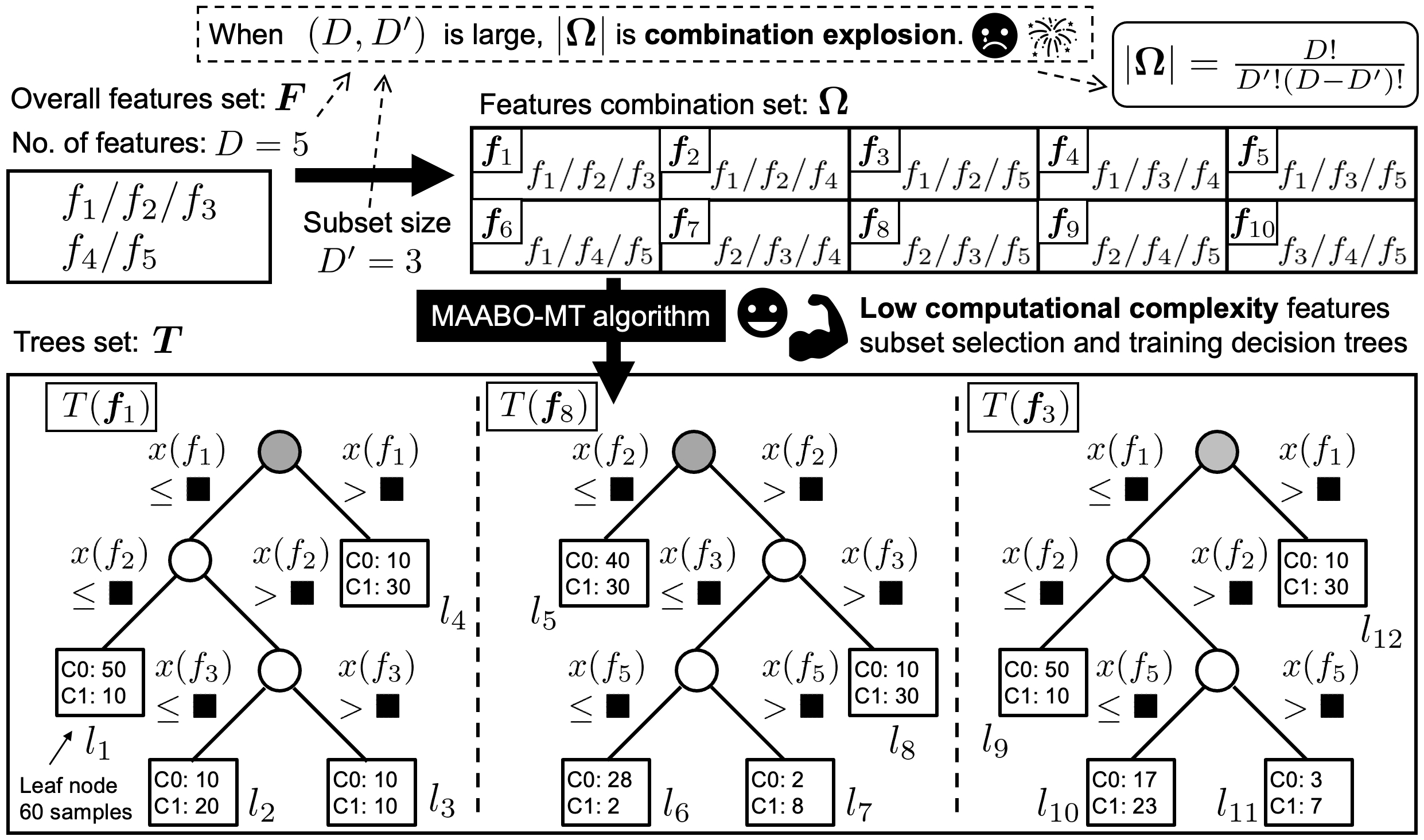}
 \caption{
Input--output relationship of MAABO-MT.
Here, the FCS $\bm{\Omega} = \{ \bm{f}_1, \cdots, \bm{f}_{10} \}$ is constructed by considering three out of five features $\bm{F} = \{f_1, \cdots, f_5\}$, where $\bm{f}_1 = \{f_1, f_2, f_3\}, \cdots, \bm{f}_{10} = \{f_3, f_4, f_5\}$.
Thereafter, only the three decision trees $\bm{T} = \{T(\bm{f}_1), T(\bm{f}_3), T(\bm{f}_8)\}$ that are expected to have high estimation performance are constructed.
Notably, if the number of features $D$ is large, the size of the FCS $|\bm{\Omega}|$ is assumed to explode and the construction of all trees will not be possible (see Equation (\ref{eq_comb_size})). Therefore, MAABO-MT is used to solve this problem.}
 \label{f_mta}
\end{figure*}

This paper proposes an algorithm to solve the aforementioned existing issues.
The input--output relationship is shown in Fig. \ref{f_mta}.
Three out of five features $f_1, \cdots, f_5$ are considered to construct decision trees.
Since three out of five give ten combinations, we define $\bm{f}_1, \cdots, \bm{f}_{10}$.
Ten decision trees can be constructed, but some trees may not perform satisfactorily.
Therefore, the example in Fig. \ref{f_mta} uses only three constructed decision trees that are expected to perform well.
On a small scale, all ten decision trees can be constructed without searching.
However, doing so is impossible when the number of features exceeds a certain threshold.
A set comprising $D$ features can be defined as
\begin{align}
\bm{F} = \{f_1, \cdots, f_{D} \}, \label{eq_fe}
\end{align}
where $f_i \in \bm{F}$ represents a feature identifier such as ``age.''
Vectors and matrices did not appear in this study, but they do exist in several other sets; therefore, bold font is used to denote sets.
Let $\bm{f}$ be the feature subset obtained by extracting $D^\p$ features from an overall feature set $\bm{F}$.
Then, the features combination set (FCS) is defined by
\begin{align}
\bm{\Omega} = \{ \bm{f} \mid \bm{f} \subset \bm{F} \land |\bm{f}| = D^\p \},\ D > D^\p, \label{eq_comb}
\end{align}
where the element $\bm{f}$ of $\bm{\Omega}$ is an unordered set.
Particularly, when $D^\p=3$, $\bm{f} \in \bm{\Omega}$ is a feature combination such that $\bm{f}_1 = \{f_1, f_2, f_3\}$ and $\bm{f}_2 = \{f_1, f_2, f_4\}$.
Additionally, $\{f_1, f_2, f_3\}$ and $\{f_3, f_2, f_1\}$ are the same elements and are not treated separately because $\bm{f}$ is an unordered set.
Therefore, when $(D, D^\p) = (5, 3)$, FCS becomes $\{ \bm{f}_1, \cdots, \bm{f}_{10} \} \in \bm{\Omega}$, as shown in the upper right section of Fig. \ref{f_mta}.
Since $\bm{\Omega}$ comprises subsets obtained by extracting $D^\p$ features from $D$ features, its size is given by
\begin{align}
|\bm{\Omega}| = {}_D C_{D^\p} = \frac{D!}{D^\p! (D - D^\p)!}. \label{eq_comb_size}
\end{align}
In Fig. \ref{f_mta}, the size is $|\bm{\Omega}| = 5!/(5-3)! = 10$ as $(D, D^\p) = (5, 3)$.
However, when $D$ is large, the construction of all decision trees become challenging.
For example, when $(D, D^\p) = (100, 5)$, the size is $|\bm{\Omega}| \simeq 7 \times 10^7$.
Therefore, an efficient search for feature subsets that yield high-performance decision trees from $\bm{\Omega}$ is necessary.

The proposed algorithm uses Bayesian optimization \cite{tpe_origin1} for searching high-performance solutions using non-parametric probability distribution.
Particularly, Gaussian and Aitchison-Aitken (AA kernel) kernels are used for solutions comprising real numbers and categorical values, respectively \cite{def_tpe1, tpe_aa1}.
Nevertheless, since this study focuses feature subset search, Bayesian optimization cannot be applied using the aforementioned kernel functions.
Therefore, we propose a framework for direction application of Bayesian optimization to subset search, by using a new modified function of the AA kernel ``modified AA (MAA).''
Thus, the proposed algorithm is called ``MAA function-based Bayesian optimization for making trees (MAABO-MT).''

\begin{figure*}[t]
 \centering
 \includegraphics[scale=0.19]{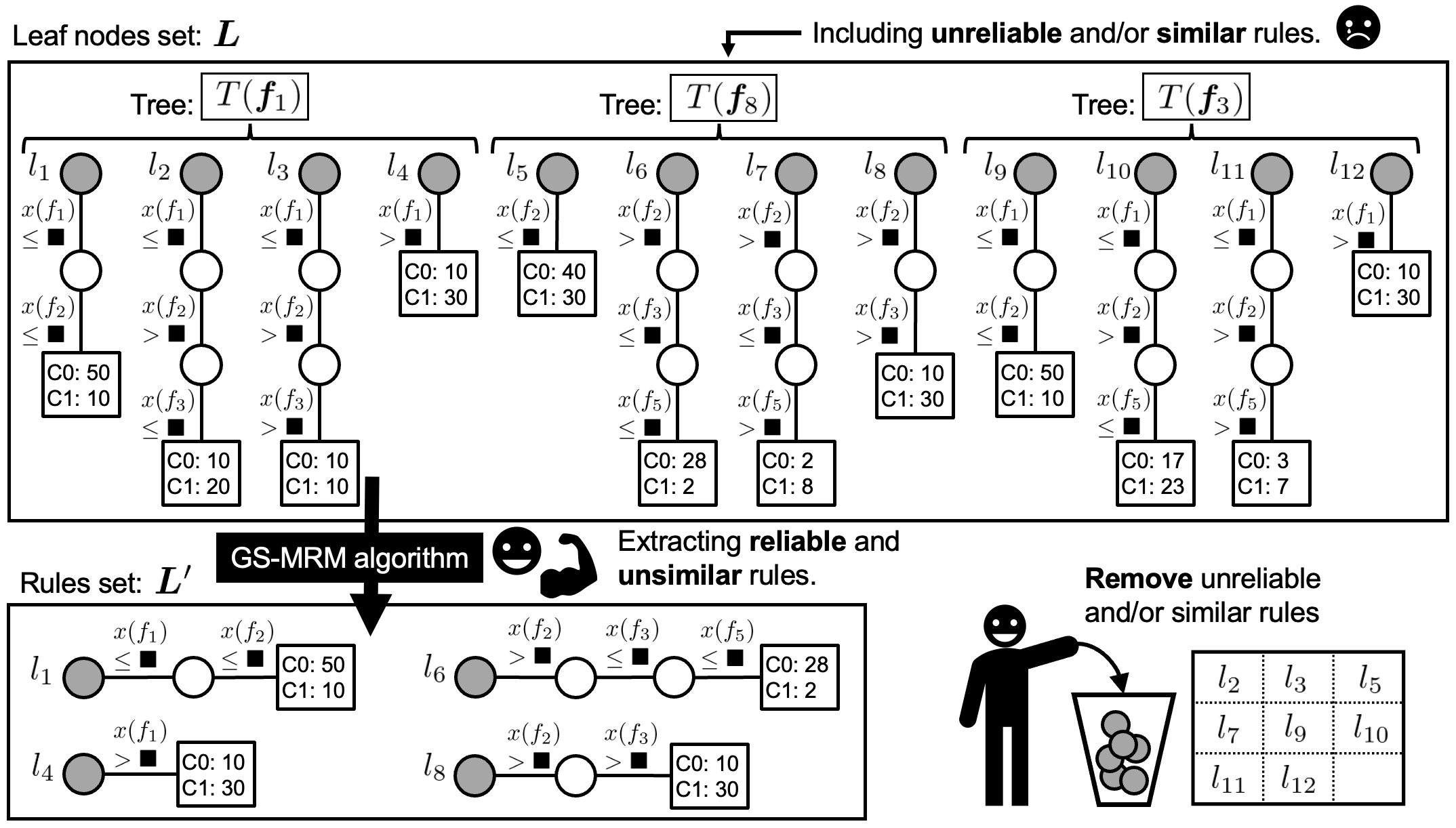}
 \caption{
Input--output relationship of GS-MRM.
Among the total 12 leaf nodes obtained from the three decision trees shown in Fig. \ref{f_mta}, only reliable and dissimilar rules are extracted.
}
 \label{f_mrm}
\end{figure*}

At a low computational cost, high-performance decision trees are constructed using the proposed method; however, only appropriate rules should be extracted because some leaf nodes have small sample sizes, whereas others do not have clearly separated classes.
The number of reliable leaf nodes is not numerous, some of which are similar, thereby making proper academic discussion from all leaf nodes difficult.
Therefore, we propose an algorithm to extract leaf nodes such that the following conditions are satisfied. (1) The sample size is sufficient, (2) classes are clearly divided, and (3) leaf nodes are not similar to the previously extracted ones.
The input--output relationship of this algorithm is shown in Fig. \ref{f_mrm}.
Herein, four leaf nodes are extracted from a total of 12 leaf nodes $l_1, \cdots, l_{12}$.
C0 and C1 represent the class labels and the numbers in the square boxes represent the sample sizes.
Initially, $l_7$ and $l_{11}$ are automatically removed owing to their small sample size. Thereafter, $l_5$ and $l_{10}$ are automatically removed as no clearly separated classes exist.
Additionally, for similar rules such as $l_1$ and $l_9$, only one rule can be adopted.
Therefore, the removals result in an output of only trusted leaf nodes ($l_1, l_4, l_6, l_8$).
Since the method combines Gini index \cite{def_gini} and Simpson coefficient \cite{def_simp}, we termed it as ``Gini and Simpson coefficients-based multi-rules mining algorithm (GS-MRM algorithm).'' This paper describes the proposed algorithms in detail.

\section{Related works on decision tree}
Decision trees are one of the most commonly used data-mining methods \cite{top10x}, with several proposed algorithms such as ID3 \cite{tree_quinlan1}, C4.5 \cite{c45fast}, CHAID \cite{chaid_tree1}, and CART \cite{cart1, CART2}, which use all features to construct a single decision tree.
Decision trees can improve the estimation performance through ensemble learning.
For example, the approach of constructing multiple decision trees using bootstrapping called RF \cite{rf_origin1} has been applied in various domains \cite{rf_varid1, rf_varid2, rf_varid3} including feature importance measurements \cite{fe_imp_rf1, oob_fe_li_2020}.
Additionally, some studies have analyzed the internal structure of RF \cite{rf_ex1, rf_ex2, rf_ex3}.

A gradient boosting decision tree \cite{gb_tree1}, such as XGboost \cite{xg_origin} and LightGBM \cite{lightbgm_origin}, has been proposed using ensemble learning.
Compared to random approaches, the aforementioned methods are more strategic in terms of error reduction and expected to perform better than RF.
Huang et al. \cite{xg_appl1} and Joharestani et al. \cite{xg_appl2} reported a better performance of XG-boost relative to RF.
Additionally, some studies reported a superior estimation performance of LightGBM than that of RF \cite{lightgbm_app1, lightgbm_app2}.

As aforementioned, decision trees are evolving and the development of ensemble learning is particularly remarkable. Since previous studies mainly focused on improving estimation performance, our study focuses on rule mining and not on performance estimation. We propose an algorithm that discovers several rules latent in data at a small computational cost.

\section{Proposed method}
\subsection{MAABO-MT algorithm} \label{subsec_pema2}
The MAABO-MT algorithm shown in Fig. \ref{f_mta} is a partial modification of Bayesian optimization \cite{tpe_origin1, def_tpe1}, an algorithm for extracting feature subsets from FCS $\bm{\Omega}$ that leads to high performance decision trees.
Initially, the following sets are defined.
\begin{itemize}
\item Unverified features combination set (U-FCS): $\bm{F}^\p = \bm{\Omega}$
\item Verified features combination set (V-FCS): $\ov{\bm{F}^\p} = \varnothing$
\item Trees set: $\bm{T} = \varnothing$
\end{itemize}
Herein, $\bm{F}^\p$ is a set comprising feature subsets for which estimation performance has not been verified and is initialized by $\bm{\Omega}$, while $\ov{\bm{F}^\p}$ represents a set comprising feature subsets where the estimation performance has been verified and initialized with the empty set $\varnothing$.
Since MAABO-MT is an algorithm that constructs decision trees, the decision tree set is initialized with an empty set $\bm{T} = \varnothing$.

Initially, we randomly construct $N_\m{I}$ trees.
Based on these validation performances, the next step involves feature exploration for tree construction.
Therefore, $N_\m{I}$ denotes the initial solution size.
Training data are used to construct decision trees, validation data are used to measure and tune the validation performance and hyperparameter (maximum depth of trees), respectively, and no overlap exists between the two datasets.

First, the generation and evaluation of initial solutions are described.
A feature subset $\bm{f}$ is randomly selected from U-FCS $\bm{F}^\p$, using which the maximum tree depth that leads to maximum verification performance is determined as 
\begin{align}
p^* = \argmax_{p \in \{1, \cdots, p_\m{max}\} } S(\bm{f}, p), \label{eq_search_dep}
\end{align}
where $S(\bm{f}, p)$ is the validation performance of the decision tree constructed by maximum depth $p$, feature subset $\bm{f}$, and the CART algorithm \cite{CART2}.
$p_\m{max}$ denotes the optimal maximum depth.
The performance index is the macro-average F1 score in the classification task.
The aforementioned process is performed to prevent overfitting.
Using the optimal maximum depth $p^*$ and a feature subset $\bm{f}$, we construct a decision tree $T(\bm{f}, p^*)$ that is added to the decision tree set as 
\begin{align}
\bm{T} = \bm{T} \cup \{ T(\bm{f}, p^*) \}. \label{eq_up_tree}
\end{align}
Since the aforementioned procedure validates the feature subset $\bm{f}$, V-FCS and U-FCS are updated as
\begin{align}
\ov{\bm{F}^\p} = \ov{\bm{F}^\p} \cup \{\bm{f}\}, \ \ \bm{F}^\p = \bm{F}^\p \setminus \{\bm{f}\}. \label{eq_up_fp_eq_up_fov}
\end{align}
By implementing the procedure $N_\m{I}$ times, a set $\bm{T}$ comprising $N_\m{I}$ decision trees is constructed.
Therefore, the initial solution is generated following the aforementioned procedure.

In V-FCS $\ov{\bm{F}^\p}$, some feature subsets have a high validation performance, while others have a low performance.
For clarification, V-FCS $\ov{\bm{F}^\p}$ is sorted in descending order of verification performance, making it an ordered set, such that 
\begin{align}
\ov{\bm{F}^\p} = \{ \ov{\bm{f}_1}, \ov{\bm{f}_2}, \cdots \}^*, \ \ \ov{\bm{f}_1} \succcurlyeq \ov{\bm{f}_2} \succcurlyeq \cdots,  \label{eq_srt_set}
\end{align}
where ``$\succcurlyeq$'' represents an ordered relationship based on verification performance and $\{ \cdot \}^*$ represents an ordered set.
Then, let $\bm{U}^{+}$ and $\bm{U}^{-}$ denote the sets comprising the upper and lower elements of $\ov{\bm{F}^\p}$, respectively, such that 
\begin{align}
\bm{U}^+ &= \{\ov{\bm{f}_n} \ | \ n = 1, \cdots, N_\m{th}\}, \nonumber \\
\bm{U}^- &= \{\ov{\bm{f}_n} \ | \ n = N_\m{th} + 1, \cdots, |\ov{\bm{F}^\p}| \}, \nonumber \\
\ov{\bm{F}^\p} &= \bm{U}^+ \cup \bm{U}^-, \ \ N_\m{th} = \lfloor \alpha |\ov{\bm{F}^\p}| \rfloor, \ \ \alpha \in (0, 1)
\end{align}
where $\alpha$ denotes the split coefficient.
$\bm{U}^+$ and $\bm{U}^-$ are known as high and low-score feature combination sets (H-FCS and L-FCS), respectively.
Using $\bm{U}^+$ and $\bm{U}^-$, the next feature subset to be validated is searched from U-FCS $\bm{F}^\p$.
However, the size of U-FCS $|\bm{F}^\p|$ explodes when $D$ and $D^\p$ are large, as shown in (\ref{eq_comb_size}).
Therefore, the search space should be reduced.

\begin{figure}[t]
 \centering
 \includegraphics[scale=0.6]{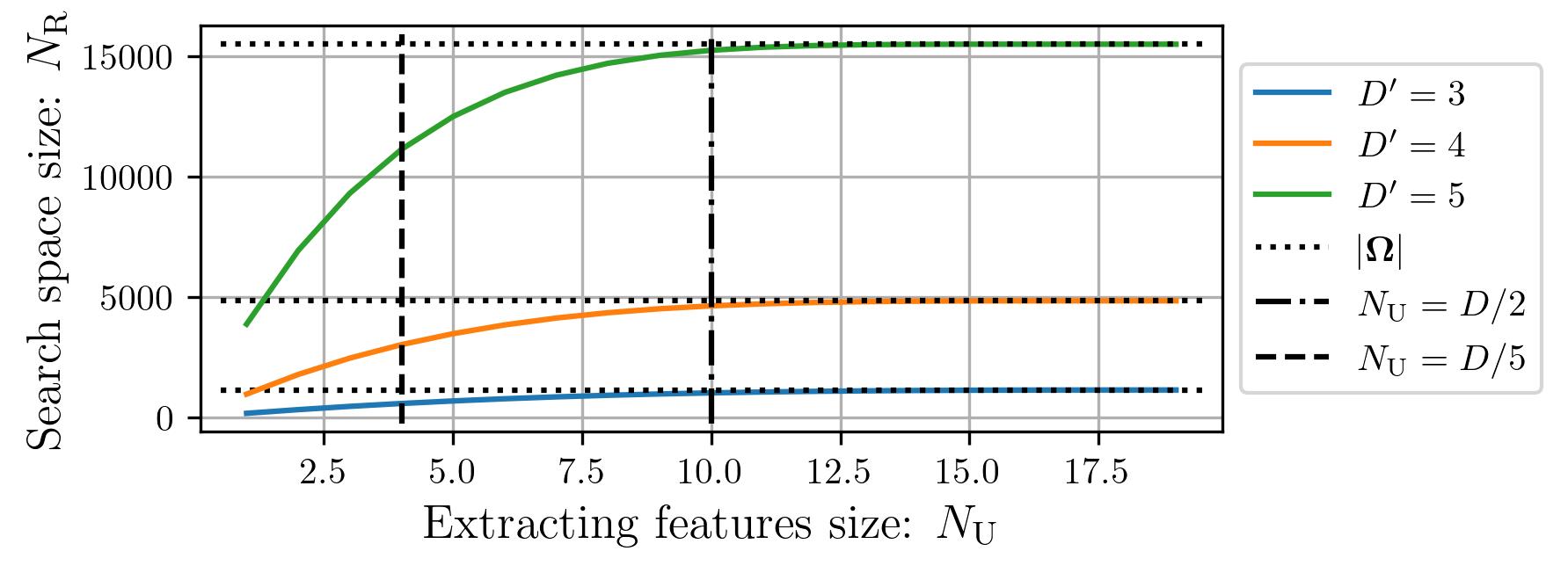}
 \caption{Relationships between extracting features $N_\m{U}$ and search space size $N_\m{R}$.}
 \label{f_nq}
\end{figure}

The individual features used in each subset within $\bm{U}^+$ can contribute to higher performance.
Therefore, the number of individual features $f_1, \cdots, f_D \in \bm{F}$ present in $\bm{U}^+$ are counted, and the top $N_\m{U}$ features that are most frequently used are extracted.
Subsequently, the subsets in U-FCS $\bm{F}^\p$ using at least one of the extracted features are chosen as a search space.
Herein, the number of unselected features is denoted as $D-N_\m{U}$.
Therefore, the search space is reduced by the number of feature subsets obtained by extracting $D^\p$ features from $D-N_\m{U}$ features.
Therefore, the reduction in the search space size is given by ${}_{D-N_\m{U}}C_{D^\p}$, and the reduced search space size is expressed as
\begin{align}
N_\m{R} = |\bm{\Omega}| - {}_{D-N_\m{U}}C_{D^\p}. \label{eq_n_q}
\end{align}
Since ${}_{x}C_{y} = 0 \ (x < y)$, $N_\m{U} \le D - D^\p$ is required to reduce the search space.
The relationship between $N_\m{U}$ and $N_\m{R}$ is shown in Fig. \ref{f_nq}.
The obtained result $N_\m{R} < |\bm{\Omega}|$ indicates the reduction in search space.
Particularly, when $N_\m{U} = D/5$, the search space is reduced by approximately half.

Smaller values of $N_\m{U}$ reduce the search space; however, extremely small values can result in a localized search.
Therefore, setting $N_\m{U}$ to an extremely small value is undesirable.
To perform search at a low computational cost when $N_\m{U}$ is not too small, we randomly select $N_\m{E}$ feature subsets from $N_\m{R}$ candidates.
Then, the next feature subset to be verified is selected from among the $N_\m{E}$ subsets.
In other words, two parameters reduce the search space in this algorithm including: $N_\m{U}$ and $N_\m{E}$.
Therefore, search space obtained using the aforementioned procedure can be defined as
\begin{align}
\bm{F}^{\p \p} = H(\bm{F}^\p, \bm{U}^+, N_\m{U}, N_\m{E}). \label{eq_n_q_n_e}
\end{align}
where $H$ is a function that takes the top $N_\m{U}$ single features that are frequently present in $\bm{U}^+$ and randomly selects $N_\m{E}$ subsets containing the top features from $\bm{F}^\p$.
The smaller the size of $N_\m{E}$, the lower is the computational cost.
However, if $N_\m{E}$ is too small, the subset to be adopted depends on random luck.

Next, we consider the feature subset selection from $\bm{F}^{\p \p}$ to construct a high-performance decision tree.
Assuming two distributions to estimate the probabilities that $\bm{f} \in \bm{F}^{\p \p}$ belongs to $\bm{U}^+$ and $\bm{U}^-$, we have 
\begin{align}
p(\bm{f} |\bm{U}^+), \ p(\bm{f} | \bm{U}^-), \  \bm{f} \in \bm{F}^{\p \p}. \label{eq_prob_un}
\end{align}
Here, the features subset $\bm{f} \in \bm{F}^{\p \p}$ with a larger $p(\bm{f} |\bm{U}^+)$ and smaller $p(\bm{f} |\bm{U}^-)$ should be selected as the next features subset $\bm{f}^* \in \bm{F}^{\p \p}$, such that
\begin{align}
\bm{f}^* = \argmax_{\bm{f} \in \bm{F}^{\p \p}}\frac{p(\bm{f}|\bm{U}^+)}{p(\bm{f}|\bm{U}^-)}. \label{eq_argmax_f}
\end{align}
Then, using the feature subset $\bm{f}^*$, \ref{eq_search_dep}, \ref{eq_up_tree}, and \ref{eq_up_fp_eq_up_fov} are processed, while the sets $\bm{T}, \ov{\bm{F}^\p}, \bm{F}^\p$ are updated.
When the aforementioned process is repeated iteratively, high-performance trees are strategically constructed.
If the number of iterations is $N_\m{B}$, the final number of trees constructed is $N_\m{B}+N_\m{I}$, using $N_\m{I}$ number of initial solutions.
The procedures discussed thus far form MAABO-MT, and the exact procedure is described in Algorithm \ref{alg1}.

\begin{figure}[!t]
\begin{algorithm}[H]
  \caption{MAABO-MT algorithm}
  \label{alg1}
  \begin{algorithmic}[1]
  \REQUIRE Overall features set $\bm{F}$, 
  features subset size $D^\p$, 
  initial solution size $N_\m{I}$, 
  split coefficient $\alpha$,
  iteration of Bayesian optimization $N_\m{B}$,
  maximum tree's depth $p_\m{max}$,
  distribution degree to mismatches $h$,
  damping coefficient $b$,  
  extracting single feature size $N_\m{U}$,
  sampling size $N_\m{E}$
  \ENSURE Trees set $\bm{T}$
  \STATE Creating FCS $\bm{\Omega}$ based on $D^\p$ and $\bm{F}$
  \STATE Creating U-FCS: $\bm{F}^\p \leftarrow \bm{\Omega}$
  \STATE Initializing V-FCS: $\ov{\bm{F}^\p} \leftarrow \varnothing$
  \STATE Initializing trees set: $\bm{T} \leftarrow \varnothing$
  \FOR{$n = 1$ \TO $N_\m{I}$} 
  \STATE Randomly selecting a features subset: $\bm{f} \in \bm{F}^\p$
  \STATE Optimal depth: $p^* \leftarrow \argmax_{p \in \{1, \cdots, p_\m{max}\} } S(\bm{f}, p)$
  \STATE Updating trees set: $\bm{T} \leftarrow \bm{T} \cup \{ T(\bm{f}, p^*) \}$
  \STATE Updating V-FCS: $\ov{\bm{F}^\p} \leftarrow \ov{\bm{F}^\p} \cup \{\bm{f}\}$
  \STATE Updating U-FCS: $\bm{F}^\p \leftarrow \bm{F}^\p \setminus \{\bm{f}\}$
  \ENDFOR 
  \FOR{$n = 1$ \TO $N_\m{B}$}
  \STATE Descended sorting based on validation score $\ov{\bm{F}^\p}$
  \STATE Split threshold: $N_\m{th} \leftarrow \lfloor \alpha |\ov{\bm{F}^\p}| \rfloor$
  \STATE Creating H-FCS $\bm{U}^+$ and L-FCS $\bm{U}^-$ with $N_\m{th}$ and $\ov{\bm{F}^\p}$
  \STATE Sampling $\bm{F}^{\p \p} \leftarrow H(\bm{F}^\p, \bm{U}^+, N_\m{U}, N_\m{E})$
  \STATE Probability: $p(\bm{f} | \bm{U}^{i}) \leftarrow K(\bm{f}, \bm{U}^{i}, h, b),  \ i \in \{+, -\}$
  \STATE Next subset: $\bm{f}^* \leftarrow \argmax_{\bm{f} \in \bm{F}^{\p\p}}p(\bm{f}|\bm{U}^+)/p(\bm{f}|\bm{U}^-)$
  \STATE Optimal depth: $p^* = \argmax_{p \in \{1, \cdots, p_\m{max}\} } S(\bm{f}^*, p)$
  \STATE Updating trees set: $\bm{T} \leftarrow \bm{T} \cup \{ T(\bm{f}^*, p^*) \}$
  \STATE Updating V-FCS: $\ov{\bm{F}^\p} \leftarrow \ov{\bm{F}^\p} \cup \{ \bm{f}^* \}$
  \STATE Updating U-FCS: $\bm{F}^\p \leftarrow \bm{F}^\p \setminus \{\bm{f}^*\}$
  \ENDFOR
  \RETURN Trees set $\bm{T}$
  \end{algorithmic}
\end{algorithm}
\end{figure}

\subsection{Probability distribution with modified Aitchison-Aitken function}
This subsection describes the method for constructing probability distributions $p(\bm{f} |\bm{U}^{i}), i \in \{+, -\}$.
When the search target is a categorical vector, the AA kernel \cite{aa_ker_origin1} is used to construct the probability distribution.
However, the AA kernel cannot be used in our case because $\bm{f}$ is a set of features.
Therefore, this paper proposes a modified AA function (MAA).

\subsubsection{Modified Aitchison-Aitken function}
As indicated in Equation (\ref{eq_argmax_f}), the next $\bm{f}^*$ to be selected should be similar to H-FCS $\bm{U}^+$ and dissimilar to L-FCS $\bm{U}^-$.
Therefore, a function $k(\bm{f}, \bm{u})$ that measures the similarity between the two feature subsets $\bm{f}$ and $\bm{u} \in \bm{U}^{i \in \{+, -\}}$ should be created.

Since $k(\bm{f}, \bm{u})$ is used to construct the probability distributions, we aim to satisfy the following constraints:
\begin{itemize}
\item (Constraint 1) $\sum_{\bm{f} \in \bm{\Omega}} k(\bm{f}, \bm{u}) = 1, \ \forall \bm{u} \in \bm{U}^{i \in \{+, -\}}$
\item (Constraint 2) $k(\bm{f}, \bm{u}) \in [0, 1]$ 
\end{itemize}
Moreover, as $k(\bm{f}, \bm{u})$ is a function for measuring similarity, we aim to satisfy the following constraint:
\begin{itemize}
\item (Constraint 3) $D^\p - |\bm{f}_v \cap \bm{u}| > D^\p - |\bm{f}_w \cap \bm{u}| \Leftrightarrow k(\bm{f}_v, \bm{u}) < k(\bm{f}_w, \bm{u}), \ \ \bm{f}_v, \bm{f}_w \in \bm{\Omega}$
\end{itemize}
Because the feature subsets $\bm{f}$ and $\bm{u}$ are sets comprising $D^\p$ features, the number of mismatches is given by
\begin{align}
m = D^\p - |\bm{f} \cap \bm{u}|, \ \ m = 0, \cdots, D^\p. \label{eq_m}
\end{align}
Therefore, $k_0, \cdots, k_{D^\p}$ are defined as the values of $k(\bm{f}, \bm{u})$ when the number of mismatches is $m = 0, \cdots, D^\p$. Next, the values that should be assigned to $k_0, \cdots, k_{D^\p}$ to satisfy Constraints 1, 2, and 3 are considered.

First, $k(\bm{f}, \bm{u})$ based on the number of mismatches $m=0$ is defined as
\begin{align}
k_0 = 1 - h, \ h \in [0, 1]. \label{eq_k0}
\end{align}
As a similarity measure between $\bm{f}$ and $\bm{u}$, $k_0$ must be large when the number of mismatches $m=0$.
Therefore, $k_0 = 1$ is not possible from Constraint 1 because values must also be assigned for cases where the number of mismatches $m = 1, \cdots, D^\p$.
Therefore, the distribution amount for $m \ge 1$ is $h$.

Next, the usage of the distributed $h$ is considered depending on the number of feature subsets with one or more mismatches in $\bm{\Omega}$.
Therefore, the number of feature subsets that include $m$ mismatches is defined as $a_m$.
From Constraint 1, $k_m$ must satisfy 
\begin{align}
a_0 k_0 + a_1 k_1 + \cdots +  a_{D^\p} k_{D^\p} = 1. \label{eq_kall}
\end{align}
Here, $a_0 = 1$ because $m$ is zero only when $\bm{f}$ and $\bm{u}$ are the same.
Additionally, by substituting Equation (\ref{eq_k0}) into Equation (\ref{eq_kall}), we obtain
\begin{align}
a_1 k_1 + \cdots +  a_{D^\p} k_{D^\p} = h. \label{eq_h}
\end{align}
Next, to satisfy Constraint 3, 
\begin{align}
k_m = b^{m-1} k_1, \ b \in (0, 1), \ m = 1, \cdots, D^\p \label{eq_km1}
\end{align}
is set up, where $b$ is the damping coefficient.
Since $k_m$ is the value of $k_1$ damped by $b^{m-1}$, when $b \in (0, 1)$, $k_1 > k_2 > \cdots > k_{D^\p}$ is satisfied.
Additionally, if $k_1$ is clarified, the $k_2, \cdots, k_{D^\p}$ can be determined.
Therefore, to clarify $k_1$, Equation (\ref{eq_km1}) is substituted into Equation (\ref{eq_h}) to obtain 
\begin{align}
k_1 = \frac{h}{a_1 b^0 + \cdots +  a_{D^\p} b^{D^\p-1}}. \label{}
\end{align}
By substituting the results into Equation (\ref{eq_km1}), we have
\begin{align}
k_m = \frac{b^{m-1} h}{\sum_{i=1}^{D^\p} a_i b^{i-1}}, \ m = 1, \cdots, D^\p. \label{eq_km}
\end{align}
As $h$ and $b$ are hyper-parameters, the unknown variable is only $a_i$.
Since $a_i$ is the total number of feature subsets with $i$ mismatches between $\bm{u}$ and $\bm{f} \in \bm{\Omega}$, its values is obtained by multiplying the following two terms,
\begin{itemize}
\item (1) the number of combinations obtained by selecting $D^\p - i$ features from among the features contained in $\bm{u}$
\item (2) the number of combinations obtained by selecting $i$ features from the features not included in $\bm{u}$
\end{itemize}
As $\bm{u}$ comprises $D^\p$ features, term (1) is ${}_{D^\p}C_{D^\p - i}$.
Since the number of features not included in $\bm{u}$ is $D - D^\p$, term (2) is denoted as ${}_{D-D^\p}C_{i}$.
Therefore, $a_i$ is obtained as 
\begin{align}
a_i &= ({}_{D^\p} C_{D^\p - i}) ({}_{D-D^\p} C_{i}), \label{eq_a}
\end{align}
where ${}_{x} C_{0} = 1$ and ${}_{x} C_{y} = 0, (x < y)$.
Upon substituting into Equation (\ref{eq_km}), we obtain
\begin{align}
k_m &= \frac{b^{m-1} h}{\sum_{i=1}^{D^\p} ({}_{D^\p} C_{D^\p - i}) ({}_{D-D^\p} C_{i}) b^{i-1}}, m = 1, \cdots, D^\p. \label{eq_km2}
\end{align}
Thus, the specific values of $k_0, \cdots, k_{D^\p}$ are determined.

Based on Equations (\ref{eq_m}), (\ref{eq_k0}), and (\ref{eq_km2}), we propose 
\begin{align}
&k(\bm{f}, \bm{u}, h, b) = \nonumber \\
&\left\{
\begin{array}{ll}
1-h, & \text{if } D^\p - |\bm{f} \cap \bm{u}| = 0\\
{\displaystyle \frac{b^{D^\p - |\bm{f} \cap \bm{u}|-1}}{\sum_{i=1}^{D^\prime} ({}_{D^\p} C_{D^\p - i}) ({}_{D-D^\p} C_{i}) b^{i-1}} h ,} &  \text{if } D^\p - |\bm{f} \cap \bm{u}| > 0
\end{array},
\right. \nonumber \\
&h \in [0, 1], \ \ b \in (0, 1),
 \label{eq_ma2}
\end{align}
as the similarity function between feature subsets $\bm{f}$ and $\bm{u}$.
The function is designed satisfy the Constraints 1, 2, and 3 and is therefore applicable as a discrete probability distribution.
$k(\bm{f}, \bm{u}, h, b)$ is a proposed function called MAA.
The design for distributing $1-h$ in the case of a match, and $h$ in the case of a mismatch also appears in the AA kernel \cite{aa_ker_origin1}.
The relationships between $k(\bm{f}, \bm{u}, h, b)$, $m$, $a_m$ are discussed in Appendix 2.

\subsubsection{Probability distribution}
This subsection describes the process of formulating the probability $p(\bm{f}|\bm{U}^{i})$, where a feature subset $\bm{f} \in \bm{\Omega}$ belongs to $\bm{U}^{i}$, using MAA function $k(\bm{f}, \bm{u}, h, b)$.

As the MAA function is calculated for a single $\bm{u} \in \bm{U}^{i}$, $k(\bm{f}, \bm{u})$ is calculated for all $\bm{u} \in \bm{U}^{i}$ and averaged as 
\begin{align}
K(\bm{f}, \bm{U}^{i}, h, b) = \frac{1}{|\bm{U}^{i}|}\sum_{\bm{u} \in \bm{U}^{i}} k(\bm{f}, \bm{u}, h, b), \ i \in \{+, -\}.
\end{align}
The MAA function satisfies the definition of probability distribution for the input $\bm{f} \in \bm{\Omega}$ because the probability of occurrence of all events is 1, such that 
\begin{align}
P(\bm{\Omega}) &= \sum_{\bm{f} \in \bm{\Omega}} K(\bm{f}, \bm{U}^{i}) \nonumber \\
&=  \frac{1}{|\bm{U}^{i}|}\sum_{\bm{u} \in \bm{U}^{i}} \sum_{\bm{f} \in \bm{\Omega}} k(\bm{f}, \bm{u}) \nonumber \\
&=  \frac{1}{|\bm{U}^{i}|}\sum_{\bm{u} \in \bm{U}^{i}} 1 \nonumber \\
& = 1, \ \ \because \sum_{\bm{f} \in \bm{\Omega}} k(\bm{f}, \bm{u}) = 1.
\end{align}
Additionally, 
\begin{align}
K(\bm{f}, \bm{U}^{i}) \in [0, 1], \ \ \forall \bm{f}, \ \ \because k(\bm{f}, \bm{u}) \in [0, 1]
\end{align}
is satisfied.
Therefore, $K(\bm{f}, \bm{U}^{i})$ denotes a discrete probability distribution.
Based on the aforementioned discussion, we adopt $K(\bm{f}, \bm{U}^{i}, h, b)$ as $p(\bm{f} | \bm{U}^{i})$, such that
\begin{align}
p(\bm{f} | \bm{U}^{i}) = K(\bm{f}, \bm{U}^{i}, h, b),  \ i \in \{+, -\}. \label{eq_prob_dist_q}
\end{align}
Consequently, Equation (\ref{eq_argmax_f}) can be calculated, and MAABO-MT can be performed.

\subsection{GS-MRM algorithm} \label{subsec_gs}
The MAABO-MT algorithm can be used to construct $N_\m{I}+N_\m{B}$ decision trees that are expected to exhibit a high verification performance.
However, when considering a rule-mining algorithm, returning all leaf nodes as output to the user is inappropriate because only a limited number of leaf nodes belonging to the decision tree are reliable and some of the multiple leaf nodes are similar to each other.
Therefore, we propose the GS-MRM algorithm as a method for extracting reliable and non-similar leaf nodes from a large number of leaf nodes of $N_\m{I}+N_\m{B}$ decision trees.

Initially, the set $\bm{L}$ comprising $N_\m{L}$ leaf nodes in $N_\m{I}+N_\m{B}$ decision trees and the set $\bm{L}^\p$ comprising the leaf nodes that are provided as outputs to the users are defined as
\begin{align}
\bm{L} = \{l_n \mid n = 1, \cdots, N_\m{L} \}, \ \bm{L}^\p = \varnothing.
\end{align}
Herein, $\bm{L}^\p$ is initialized with an empty set because the leaf nodes to be returned have not yet been determined.
The method for moving leaf nodes in $\bm{L}$ to $\bm{L}^\p$ is defined as GS-MRM.

\begin{figure}[!t]
\begin{algorithm}[H]
  \caption{GS-MRM algorithm}
  \label{alg2}
  \begin{algorithmic}[1]
  \REQUIRE Trees set $\bm{T}$, 
  threshold of leaf node samples $\beta$,
  threshold of Gini coefficient $\gamma$,
  threshold of Simpson coefficient $\delta$,
  class label size $C$
  \ENSURE Rules set $\bm{L}^\p$
  \STATE Creating leaf nodes set $\bm{L}$ based on the trees set $\bm{T}$
  \STATE Initializing rules set: $\bm{L}^\p \leftarrow \varnothing$
  \STATE Initial leaf node size: $N_\m{L} \leftarrow |\bm{L}|$
  \FOR{$i=1$ \TO $N_\m{L}$}
  \IF{$g(l_i) \ge \gamma \left( 1 - \frac{1}{C} \right)$ \OR $n(l_i) < \beta$}
  \STATE Removing leaf node: $\bm{L} \leftarrow \bm{L} \setminus \{l_i\}$
  \ENDIF
  \ENDFOR
  \WHILE{$\bm{L} \neq \varnothing$}
  \IF{$\bm{L}^\p = \varnothing$}
  \STATE Opt. leaf node: $l^* \leftarrow \argmin_{l \in \bm{L}} g(l)$
  \ELSE
  \STATE Opt. leaf node: $l^* \leftarrow \argmin_{l \in \bm{L}} \left( g(l) + \newmax_{l^\p \in \bm{L}^\p} v(l, l^\p) \right)$
  \ENDIF
  \IF{$\max_{l^\p \in \bm{L}^\p} v(l^*, l^\p) < \delta$}
  \STATE Updating rules set: $\bm{L}^\p \leftarrow \bm{L}^\p \cup \{ l^* \}$
  \ENDIF
  \STATE Updating leaf nodes set: $\bm{L} \leftarrow \bm{L} \setminus \{ l^* \}$
  \ENDWHILE
  \RETURN $\bm{L}^\p$
  \end{algorithmic}
\end{algorithm}
\end{figure}

In this study, the moving target leaf node $l^*$ is determined as 
\begin{align}
l^* =& \argmin_{l \in \bm{L}} \left( g(l) + \max_{l^\p \in \bm{L}^\p} v(l, l^\p) \right), \nonumber \\
&\m{s.t.,} \ \ g(l) < \gamma g_\m{max}, \ \ n(l) \ge \beta,\ \  \max_{l^\p \in \bm{L}^\p} v(l, l^\p) < \delta, \nonumber \\
& \gamma \in (0, 1], \ \ \delta \in (0, 1], \label{eq_opt_leaf} 
\end{align}
where $g(l)$ is the Gini index of leaf node $l$ and $g_\m{max}$ is the maximum Gini index when the classes are evenly mixed.
The rule with a lower Gini index is adopted in preference to other rules because a leaf node with a lower Gini index is a more reliable rule.
Furthermore, only the leaf nodes below the threshold are adopted following the constraint $g(l) < \gamma g_\m{max}$.
If the total sample size of a leaf node is $N$, the sample size of class $c_i$ is $N_i$, and the number of class labels is $C$, then, the Gini index is defined as
\begin{align}
g = 1 - \sum_{i=1}^{C} \left(\frac{N_i}{N}\right)^2. \label{eq_ggg}
\end{align}
The same expression has been expressed in Equation (1) of \cite{def_gini}.
When the Gini index is close to zero, data are clearly classified; if the index is large, the classes are mixed.
Since $N_i = N/C, \forall i$ is the most mixed state, by substituting its value into Equation (\ref{eq_ggg}), we obtain
\begin{align}
g_\m{max} = 1 - \frac{1}{C}.
\end{align} 
In the constraint, $\gamma$ is a parameter used to adjust the threshold of the Gini index. Thus, by setting $\gamma$ close to zero, only leaf nodes with clearly classified data are extracted.
The constraint does not work properly with unbalanced class data; therefore, a weighted Gini index must be used, which is calculated by giving ``balanced'' option to the class weight argument of DecisionTreeClassifier \cite{tree_scikit} in scikit-learn. Alternatively, the data can be converted into balanced data through over or under sampling \cite{imbalanced_against}.

The sample size of the leaf node $l$ denoted as $n(l)$.
Since leaf nodes comprising small samples are not reliable, we set the constraint that if the sample size is not greater than or equal to the threshold $\beta$, it cannot be adopted as a rule.
$v(l, l^\p)$ represents the similarity between the leaf node $l \in \bm{L}$ and the previously extracted leaf node $l^\p \in \bm{L}^\p$.
Owing to the presence of multiple extracted leaf nodes, the maximum similarity is calculated, and leaf nodes with smaller similarities are prioritized.
Additionally, the rules with high similarity to previously extracted ones should not be adopted.
Therefore, the constraint $\max_{l^\p \in \bm{L}^\p} v(l, l^\p) < \delta$, where $\delta$ is a parameter, is considered.
The details of $v(l, l^\p)$ are presented in Subsection \ref{sec_sim_leaf}.

The sets of $\bm{L}^\p$ and $\bm{L}$ are updated at the leaf node selected using Equation (\ref{eq_opt_leaf}), such that 
\begin{align}
\bm{L}^\p = \bm{L}^\p \cup \{l^*\}, \ \ \bm{L} = \bm{L} \setminus \{l^*\}. \label{eq_upl}
\end{align}
Only reliable and dissimilar rules are extracted by repeating Equations (\ref{eq_opt_leaf}) and (\ref{eq_upl}).
The details of GS-MRM are presented as Algorithm \ref{alg2}.

\subsection{Similarity between leaf nodes} \label{sec_sim_leaf}
To solve Equation (\ref{eq_opt_leaf}), we must design a function $v(l, l^\p)$ that measures the similarity between two leaf nodes $l$ and $l^\p$.
The representation of leaf node $l$ in the decision tree is given by
\begin{align}
l: x(f_{i1}) > \blacksquare \land x(f_{i2}) \le \blacksquare \land x(f_{i3}) > \blacksquare \land \cdots,
\end{align}
where the value of feature $f_i$ is $x(f_i)$.
In this study, a set comprising the smallest units of logic is defined as
\begin{align}
\bm{R}_l = \{f_{i1}: \m{large}, f_{i2}: \m{small}, f_{i3}: \m{large}, \cdots \}.
\end{align}
Cases where the same logic appears in a leaf node do exist.
In such cases, the logics are combined into a single logic for simplification.
First, we construct $\bm{R}_{l^\p}$ for the leaf node $l^\p$.
Thereafter, the similarity of the two sets $\bm{R}_l $ and $\bm{R}_{l^\p}$ is measured by Simpson coefficient \cite{def_simp}, such that
\begin{align}
v(l, l^\p) = \frac{|\bm{R}_l \cap \bm{R}_{l^\p}|}{\min \{|\bm{R}_l|, |\bm{R}_{l^\p}|\}}.
\end{align}
The output of aforementioned function is 1 and 0 for the most and least similarity between the sets. Thus, solutions to the optimization problem expressed in Equation (\ref{eq_opt_leaf}) are possible.

\section{Experiment 1: Effectiveness of MAABO-MT on extracting rules}
\subsection{Objective and outline} \label{subs_tit}
To verify the effectiveness of the proposed method, GS-MRM was applied to decision trees constructed using MAABO-MT for rule extraction.
The Titanic dataset often used to evaluate machine-learning performance was adopted \cite{titanic1, titanic2, titanic3}.
The data consisted of 1309 passenger details aboard Titanic, with class labels of dead (809 samples) and alive (500 samples).
The features included in this dataset were $f_1, \cdots, f_{9}$, as listed in Table \ref{tab_felabel}.
Notably, noise features are also included in Table \ref{tab_felabel}, but were not used in Experiment 1. They were used in Experiment 2 or later.
In Experiment 1, the overall feature set was $\bm{F} = \{f_1, \cdots, f_{9}\}$.
A feature subset size of $D^\p = 3$ was adopted to perform MAABO-MT, implying the selection of three features from the overall feature set comprising nine features to construct the FCS $\bm{\Omega}$.
Herein, the number of decision trees that could be constructed was equal to the size of $\bm{\Omega}$; thus, ${}_{9}C_{3}=84$ from Equation (\ref{eq_comb_size}).

To verify the effectiveness of MAABO-MT, decision trees were constructed using the following four approaches:
\begin{itemize}
\item A. {\bf All trees}: creating all 84 trees with all 84 features subsets in $\bm{\Omega}$.
\item B. {\bf MAABO-MT}: limited number of trees with features subset selected by MAABO-MT in $\bm{\Omega}$.
\item C. {\bf Randomized trees}: limited number of trees with features subset selected at random in $\bm{\Omega}$.
\item D. {\bf Single tree}: a single tree by using all 9 features.  
\end{itemize}
(A) involved the construction of all decision trees and ample computational resources were considered available.
(B) involved the construction of a limited number of decision trees using MAABO-MT, an efficient search algorithm.
(C) involved the construction of a limited number of decision trees using randomly selected feature subsets.
(D) involved the construction of a single decision tree using all prepared features.
Usually, (D) is used for the the academic discussion based on the rules obtained by a single decision tree.
In this study, (D) was prepared because a single decision tree was assumed to be insufficient for extracting a sufficient number of rules.

\begin{table}[t]
  \caption{Feature labels of Titanic dataset}
  \label{tab_felabel}
  \centering
  \begin{tabular}{cll}
    \hline
\multicolumn{1}{c}{ID} & \multicolumn{1}{c}{Feature} &\multicolumn{1}{c}{Details} \\ \hline
$f_1$ & Pclass & Ticket class. 1st (best), 2nd, 3rd (worst). \\
$f_2$ & Sex & Male or female.\\
$f_3$ & Age & passenger's age.\\
$f_4$ & SibSp & The number of siblings and spouses.\\
$f_5$ & ParCh & The number of parents and child.\\
$f_6$ & Fare & The fee of getting on the Titanic.\\
$f_7$ & Embarked C & Departure port: Cherbourg. True/False.\\
$f_8$ & Embarked Q & Departure port: Queenstown. True/False.\\
$f_9$ & Embarked S & Departure port: Southampton. True/False. \\ \hline
$f_{10}$ & Noise 1& Uniform random numbers from 0 to 1. \\
$f_{11}$ & Noise 2& Uniform random numbers from 0 to 1. \\
\multicolumn{1}{c}{$\vdots$}  & \multicolumn{1}{c}{$\vdots$} & \multicolumn{1}{c}{$\vdots$} \\
$f_{9+N_\m{noise}}$ & Noise $N_\m{noise}$& Uniform random numbers from 0 to 1. \\
    \hline
  \end{tabular}
\end{table}

The number of decision trees constructed using (B) MAABO-MT, as indicated in Algorithm \ref{alg1} was $N_\m{I} + N_\m{B}$.
We set $N_\m{I} = 10$ as the initial solution size and $N_\m{B} \in \{0, 10, \cdots, 70\}$ as the iterations of Bayesian optimization.
To investigate the relationship between the number of decision trees and performance in MAABO-MT, we set various values for $N_\m{B}$.
For comparison under equal conditions, the number of decision trees constructed using (C) was the same as in (B).
For other parameters of MAABO-MT, we adopted split coefficient $\alpha = 0.25$, maximum tree depth $p_\m{max} = 5$, distribution degree of mismatches $h = 0.5$,
damping coefficient $b = 0.5$,
extracting a single feature size $N_\m{U} = D$, and
sampling size $N_\m{E} \leftarrow \infty$.

All decision trees constructed in this study used the CART algorithm \cite{cart_id3_etc}.
To avoid overfitting, we adopted maximum depth yielding maximized the F-score macro-average of the validation dataset.
Data from 1309 samples were randomly split in a 7:3 ratio and assigned as training and validation data.
To eliminate the effects of randomness, data were split and each method was performed using 50 random seeds.
GS-MRM was used for rules extraction of the decision trees obtained using (A)--(D).
To detect reliable rules, $(\beta, \gamma, \delta) = (50, 0.3, 0.7)$ was adopted.

\begin{figure}[t]
 \centering
 \includegraphics[scale=0.70]{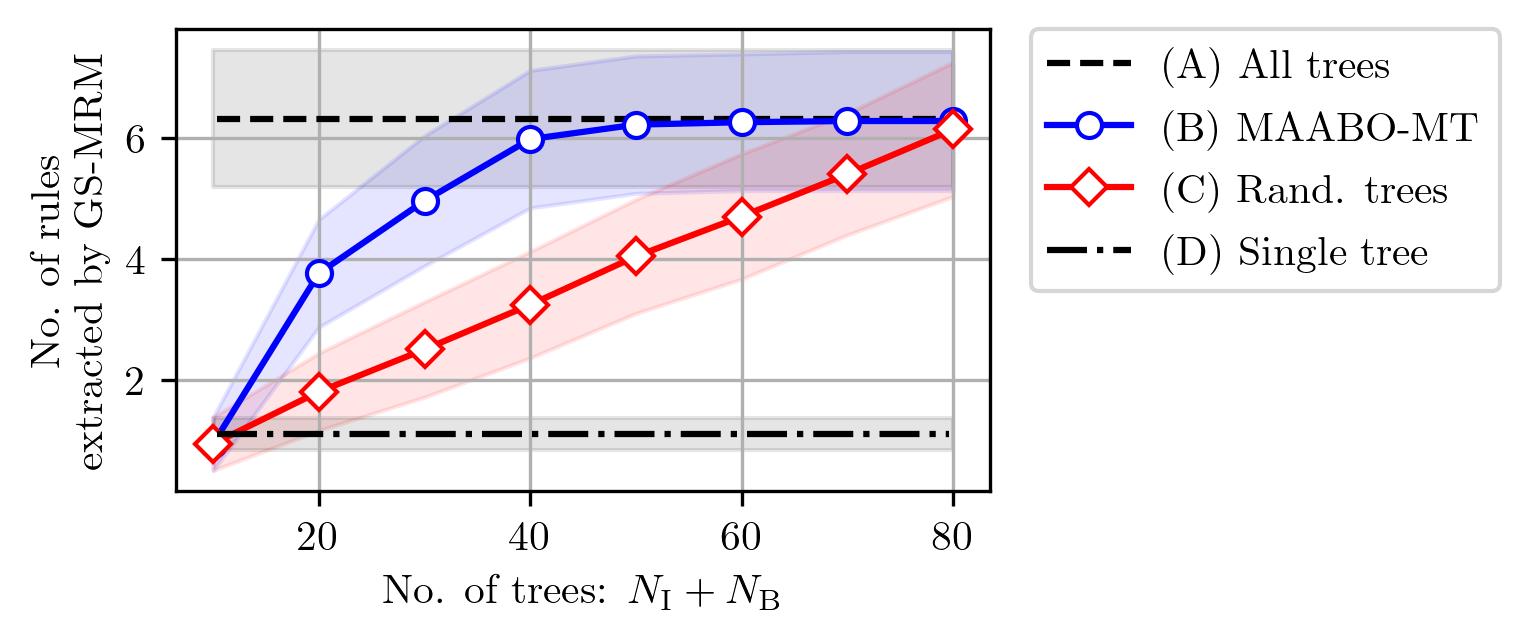}
 \caption{
 Number of extracted rules from decision trees constructed using (A)--(D) approaches.
 Each value is an average, and error bar represents 0.5 std. calculated from the results of 50 random seeds.
 }
 \label{f_ex1}
\end{figure}

\subsection{Result and discussion}
The obtained results are presented in Fig. \ref{f_ex1}.
Approximately six rules were extracted in the case of (A), while approximately one rule was extracted in the case of (D).
Thus, we confirmed that the traditional approach using all features to construct a single decision tree could only extract a fraction of the trusted rules that were latent in the data.
Contrarily, the results obtained on using (A) succeeded in extracting more rules, but ample computing resources were consumed.
This study was aimed at discovering appropriate rules without constructing all trees, as was done in (A). Therefore, the algorithm that satisfied our requirements was (B) MAABO-MT.

In the case of (B) MAABO-MT, almost all rules were found by constructing only half of all the decision trees.
Since this result was superior to that of (C) randomized trees, MAABO-MT could extract numerous reliable rules in a shorter computation time.
The details of the extracted rules are presented in Section \ref{s_ex4} and Table \ref{tab_all_fe}. Additional information are presented in Appendix 3.

\section{Experiment 2: Robustness against noise features}
\subsection{Objective and outline}
When analyzing real-world data, the dataset often contains noisy features (meaningless numerical information) without the knowledge of an analyst.
Therefore, evaluation of robustness against meaningless noise features is important.
Approaches that used all features to construct a single tree tended to adopt noisy features (Appendix 1).
Contrarily, MAABO-MT was expected to avoid the adoption of noisy features because it constructed trees by exploring feature subsets.
This section describes the results of the analysis.

Noise features generated by uniform random numbers in the Titanic dataset were added.
After $f_{10}, \cdots$ the features listed in Table \ref{tab_felabel} are the noise features, of which the number of features adopted for the experiment included $N_\m{noise} \in \{1, \cdots, 20 \}$.
Adding noise features to the nine proper features $f_1, \cdots, f_9$ included in the Titanic dataset, the total number of features $D$ ranged from 10 to 29.
The size of the FCS $|\bm{\Omega}|$ obtained by extracting the three features ($D^\p = 3$) ranged between 120 and 3,654, from Equation (\ref{eq_comb_size}).

As the setting parameters of MAABO-MT, the initial solution size $N_\m{I}$ was set to $10$ and the number of iterations of Bayesian optimization was $N_\m{B} = 100$.
Thus, 110 decision trees were constructed.
The other parameters were the same as those used in Experiment 1.
The randomized tree approach described in Experiment 1 was adopted for comparison.
The number of decision trees constructed using the randomized trees was also 110.
GS-MRM was used for rule extraction, and the adoption parameters were the same as those used in Experiment 1.
Additionally, an analysis was conducted using 50 random seeds to eliminate the effects of randomness.

\begin{figure}[t]
 \centering
 \includegraphics[scale=0.82]{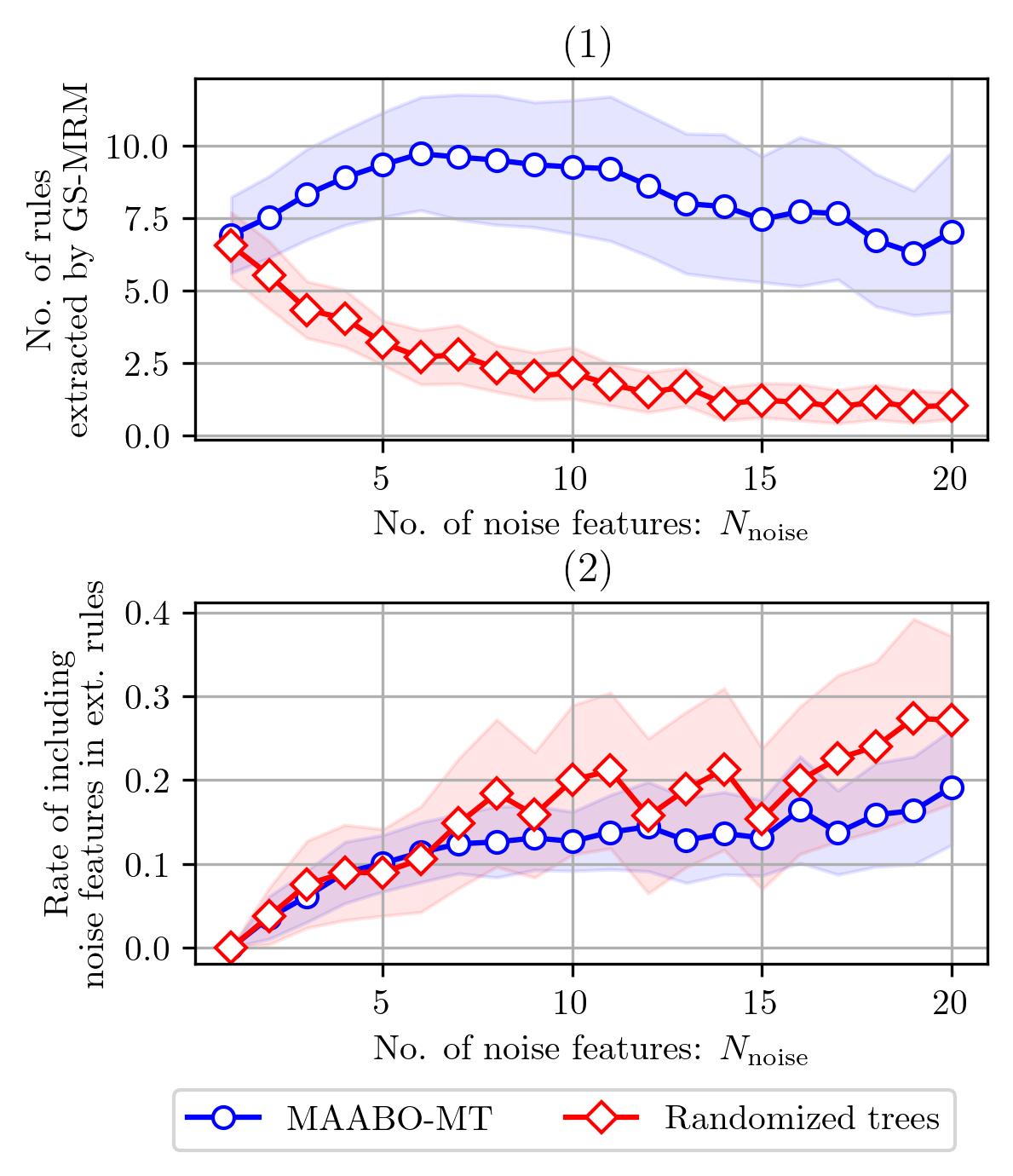}
 \caption{
 Effect on MAABO-MT search performance of noise features.
 Each value represents the average, and error bar represents 0.5 std. calculated from the results of 50 random seeds.
 }
 \label{f_ex2}
\end{figure}

\subsection{Result and discussion}
The results obtained from the aforementioned procedure are shown in Fig. \ref{f_ex2} (1), which presents the number of rules extracted by the GS-MRM.
In the case of randomized tree approach, we confirmed that the larger the number of noise features, the smaller the number of extracted reliable leaf nodes.
However, MAABO-MT confirmed that a certain number of reliable leaf nodes could be extracted even when the noise features increased. Nonetheless, cases in which noisy features were included in the extracted rules could exist.
Therefore, we calculated the noise content rate of the extracted rules,
and the results are presented in Fig. \ref{f_ex2} (2).
Thus, the noise content in the extracted rules was confirmed to increase with an increase in noise features.
Herein, MAABO-MT tended to have a slightly lower noise content than the randomized tree.
Although the difference was slight, the randomized tree could only extract a small number of rules when the noise features were large (Fig. \ref{f_ex2} (1)).
When there were more than 15 noise features, only one rule was extracted from the randomized trees.

Therefore, MAABO-MT could extract a certain number of reliable rules even when the number of noise features increased. The approach based on randomized trees was unable to extract rules when the number of noise features increased.

\section{Experiment 3: Details of rules extracted by MAABO-MT and GS-MRM algorithm} \label{s_ex4}
\subsection{Objective and outline}
This section presents the specific rules detected using MAABO-MT and GS-MRM.
The datasets used included Titanic, Boston housing \cite{dataset_boston, harrison1978hedonic}, and diabetes \cite{diadatast}.
Boston Housing is a dataset that is used to estimate home prices based on environmental factors and other attributes.
Diabetes is a dataset used to estimate disease progression based on personality and blood components.
Both datasets have been used to evaluate the performance of machine-learning algorithms \cite{tounyo1, tounyo2, boston_use1, boston_use2, Boston_use3}.
The features and class labels included in each dataset are described in Subsection \ref{aft62}.
The conditions for running MAABO-MT included the number of initial solutions $N_\m{I} = 10$ and the number of trees to be constructed is $N_\m{I} + N_\m{B} = |\Omega|/2$, it means half of $|\Omega|$.
The number of Bayesian optimization iteration is $N_\m{B} = |\Omega|/2 - N_\m{I}$.
Other parameters were the same as those used in Experiment 1.
Furthermore, we constructed a single decision tree using a traditional analysis method to compare the extracted rules.
The adopted approach used all features and constructed a single decision tree using CART.
In both approaches, the maximum depth of the tree that maximized the verification performance was explored and applied.

\subsection{Boston housing dataset} \label{aft62}

The dataset had the following features:
 $f_{1}$: CRIM: crime rate per capita,
 $f_{2}$: ZN: percentage of residential lots larger than 25,000 square feet,
$f_{3}$: INDUS: Percentage of non-retail business areas,
$f_{4}$: CHAS: Charles River binary variable (yes or no),
$f_{5}$: NOX: nitric oxides concentration,
$f_{6}$ RM: the number of rooms per residence,
$f_{7}$: AGE: percentage of old residences (before 1940),
$f_{8}$: DIS: distances to employment centers,
$f_{9}$: RAD: accessibility of highways,
$f_{10}$: TAX: property tax rate,
$f_{11}$: PTRATIO: ratio of students to teachers, and
$f_{12}$: LSTAT: low-status population. Note: Race-related features were excluded.

The total numbers of samples and prediction targets included 506 samples,
C0: cheap price (below average price, 297 samples), 
C1: high price (over-average price, 209 samples).

\subsection{Diabetes dataset}

The dataset had the following features
$f_{1}$: AGE: age in years,
$f_{2}$: SEX: male or female,
$f_{3}$: BMI: body mass index,
$f_{4}$: BP: blood pressure,
$f_{5}$: TC: total cholesterol,
$f_{6}$: LDL: low-density lipoproteins,
$f_{7}$: HDL: high-density lipoproteins,
$f_{8}$: TCH: total cholesterol,
$f_{9}$: LTG: log triglyceride level, and
$f_{10}$: GLU: glucose.

The total number of samples and prediction targets included
442 samples,
C0: low progression (average progression: 247 samples), 
C1: high progression (average progression: 195 samples).

\begin{table}[t]
  \caption{Leaf nodes size and extracted rules size}
  \label{tab_leaf_rule}
  \centering
  \begin{tabular}{cccc}
    \hline
    Method & Dataset & \begin{tabular}{c}Leaf nodes \\ size $|\bm{L}|$\end{tabular} & \begin{tabular}{c}Extracted rules \\ size $|\bm{L}^\p|$\end{tabular} \\ \hline
    MAABO-MT & Titanic & 459 & 10 \\
    &Boston & 524 & 16\\ 
    &Diabetes & 269 & 7 \\ \hline
    Single tree & Titanic & 16 & 2 \\ 
    & Boston &4& 2 \\
    & Diabetes & 8 & 1 \\    
    \hline
  \end{tabular}
\end{table}

\begin{table*}[t]
  \caption{Extracted rules by MAABO-MT and single tree (Titanic, Boston housing, and Diabetes datasets).}
  \label{tab_all_fe}
  \centering
  \begin{tabular}{ccccccl}
    \hline
    Dataset & Method & Rule ID & Class & Sample size & Gini & Extracted rule \\ \hline
    
    Titanic&MAABO-MT
    &P1& Survival & 59 & 0.00 & Sex/female, Age/high, Fare/large \\
    &&P2& Survival & 110 & 0.02 & Sex/female, Fare/large, Pclass/good \\
    &&P3& Survival & 72 & 0.04 & Sex/female, ParCh/small, Pclass/good \\
    &&P4& Death & 103 & 0.11 & Sex/male, Age/high, Pclass/bad \\
    &&P5& Death & 79 & 0.14 & Sex/male, Fare/small\\ 
    && \multicolumn{5}{c}{\#\#\# and other 5 rules \#\#\# } \\ \cline{2-7}     
    &Single tree 
    &S1& Survival & 108 & 0.01 & Sex/female, Age/high, Fare/large, Pclass/good \\
    &&S2& Death & 103 & 0.11 & Sex/male, Age/high, pclass/bad \\ \hline

    Boston & MAABO-MT
    &P1 & High price & 100 & 0.03 & LSTAT/small, RM/large\\
    &&P2 & Low price & 56 & 0.05 & AGE/large, NOX/large \\
    &&P3 & High price & 75 & 0.04 & NOX/small, RM/large \\
    &&P4 & Low price & 139 & 0.11 & PTRATIO/large, LSTAT/large \\
    &&P5 & High price & 101 & 0.11 & CRIM/small, RM/large \\ 
    && \multicolumn{5}{c}{\#\#\# and other 11 rules \#\#\# } \\ \cline{2-7}     
    & Single tree 
    &S1& Low price & 139 & 0.11 & PTRATIO/large, LSTAT/large \\
    &&S2& High price & 131 & 0.12 & RM/large, LSTAT/small \\ \hline
    
    Diabetes & MAABO-MT    
    &P1 & Low progression & 81 & 0.03 & HDL/large, BMI/small, LTG/small \\
    &&P2 & Low progression & 89 & 0.06 & AGE/small, BMI/small, LTG/small \\
    &&P3 & Low progression & 58 & 0.08 & TC/small, HDL/large, BMI/small \\
    &&P4 & High progression & 54 & 0.14 & GLU/large, BMI/large, LTG/large\\
    &&P5 & Low progression & 81 & 0.14 & AGE/small, HDL/large, BMI/small \\
    && \multicolumn{5}{c}{\#\#\# and other 2 rules \#\#\# } \\ \cline{2-7}     
    & Single tree 
    &S1& Low progression & 89 & 0.06 & AGE/small, BMI/small, LTG/small \\
    
    \hline
  \end{tabular}
\end{table*}

\subsection{Results and discussion}
The total number of leaf nodes in the constructed decision trees was represented by $|\bm{L}|$, and $|\bm{L}^\p|$ denoted the total number of rules extracted by GS-MRM, as listed in Table \ref{tab_leaf_rule}.
In the case of MAABO-MT, a large number of leaf nodes were generated, while 10, 16, and 7 rules were extracted using GS-MRM.
Conversely, in the single-tree approach, only 2, 2, and 1 reliable rules were extracted by the GS-MRM.
Although a single decision tree had several leaf nodes, only a few had a low Gini index and a large sample size.
MAABO-MT could extract several rules, whereas the traditionally used single tree could extract only a small number of rules.

The extracted rules are presented in Table \ref{tab_all_fe}.
For the Titanic dataset and single tree, the rules for survival and death were limited.
For example, S1 and S2 showed that elderly males and females were more likely to die and survive, respectively.
However, the rules extracted by MAABO-MT confirmed that females were more likely to survive (P2 and P3) regardless of age.
Thus, males were more likely to die regardless of age (P5).
If the results of only a single decision tree were considered for discussion, misunderstandings such as the age being older could arise. Such issues were eliminated by using MAABO-MT.
Additionally, MAABO-MT identified ``low ParCh (P3)'' as a survival condition, which was not identified in a single tree.
As males were more likely to die, the entire family was not in a position to escape. Therefore, single individuals were considered more likely to escape and survive.

In the case of Boston, a single tree extracted the rule (S1) ``lower housing prices when there are more students per teacher and more low-income families.''
Additionally, in the case of MAABO-MT, a different rule (P1) was extracted along with S1, such that ``If the house is old and the air is dirty, the house price is low.''
In the case of a single tree, the rule (S2) that stated ``areas with fewer low-income residents and more rooms have higher housing prices'' was extracted.
Conversely, in the case of MAABO-MT, rules such as ``areas with low crime rates have higher housing prices'' and ``areas with clean air have higher housing prices''(P3 and P5) were extracted in addition to S2.

In Diabetes and single tree case, only the rule ``low age, low BMI and low LTG, then, low progression of diabetes'' was extracted (S1).
Conversely, in the case of MAABO-MT, rules that could not be found in the single tree case, such as ``if HDL is high, diabetes is low progression'' (P1, P3, and P5) were extracted.
The results were valid because patients with diabetes had low HDL \cite{hdl1}.
Furthermore, MAABO-MT also extracted rules with high progression of diabetes that were not extracted in a single tree case (P4).
Therefore, MAABO-MT provided deeper insights than a single tree for all the datasets tested.

\section{Conclusion}
In this study, we highlight the disadvantages of rule extraction using a single decision tree, which is a traditional approach that can only discover a small fraction of the multiple rules latent in data. Therefore, we propose multi-rule mining algorithms MAABO-MT and GS-MRM to solve existing problems.
We propose an MAA function that is required to solve the feature subset search problem with Bayesian optimization.
Through several experiments using open datasets, the proposed method is found to efficiently find multiple rules with a small amount of computation.

The proposed method includes several hyperparameters. Sensitivity analysis is performed on some of the parameters, while others are not fully analyzed.
In the future, we plan to analyze all parameters to determine the recommended values for each hyperparameter.

\nocite{*}
\bibliographystyle{IEEEtran}
\bibliography{tree_bi_aims}


\section*{Appendix 1: Robustness of decision tree against noise features}
In this section, we discuss the robustness of a single decision tree to noise features using the traditional approach of CART algorithm.
For this analysis, the Titanic dataset comprising the features listed in Table \ref{tab_felabel} for 1309 samples was randomly split into training and validation data in a 7:3 ratio.
The number of noise features in the dataset varied by 1 from 0 to 20.
For hyperparameter tuning, we searched for the maximum depth and minimum sample in the leaf nodes (MSL) of the decision tree that maximized the verification performance based on macro-averaging the F-scores.
The maximum depth search range was 1--5 and MSL search range was 1--100.
Both MSL and maximum depth were parameters that affected tree complexity; however, MSL was the control variable for individual nodes, while the maximum depth was the control variable for the entire tree.
To stabilize the results, random splits of the training and validation data were performed using 100 random seeds.

\begin{figure*}[t]
 \centering
 \includegraphics[scale=0.8]{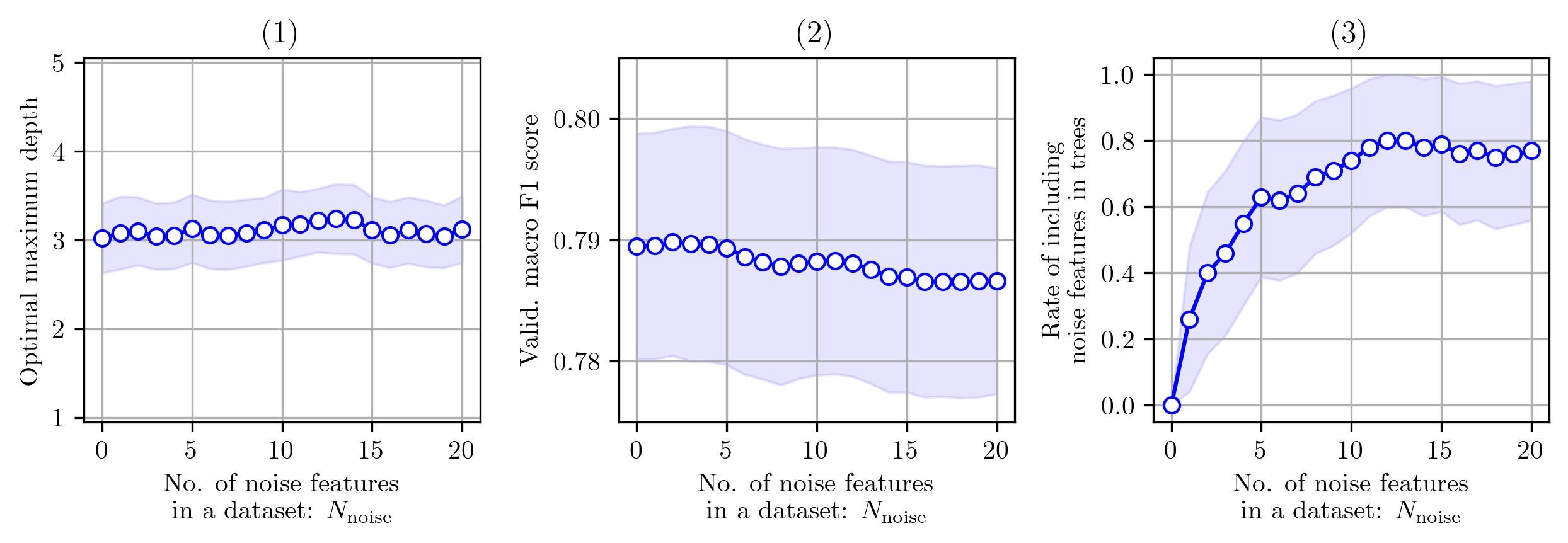}
 \caption{
 (1) Maximum depth leading to maximized validation performance, 
 (2) macro F1-scores measured from validation data, and (3) rate of including noise features in trees.
 Values in the figures are average and 0.5 std. of 100 seeds.}
 \label{f_appen}
\end{figure*}

The results are presented in Fig. \ref{f_appen}.
Fig. \ref{f_appen} (1) presents the maximum depth parameter of the decision tree explored using the validation data.
An increase in the number of noise features did not cause a clear change in the adopted maximum depth.
The depth was approximately three, which confirmed that a simple decision tree was constructed to avoid overfitting.
Fig. \ref{f_appen} (2) presents the macro-average F-scores measured using the validation data.
Increasing the number of noise features resulted in a slight decrease in the F-score, but a slight change.
Fig. \ref{f_appen} (3) represents the rate of noise features included in the decision trees constructed using 100 seeds.
The results confirmed that the greater the number of noise features in the dataset, the higher the noise content, implying that meaningless features could be adopted into the tree if they were not removed before the decision-tree construction.
Even when hyperparameter search based on the validation performance was applied, the risk existed; therefore, some type of feature selection should be performed.

\section*{Appendix 2: Numerical example of MAA function}
\begin{figure}[t]
 \centering
 \includegraphics[scale=0.8]{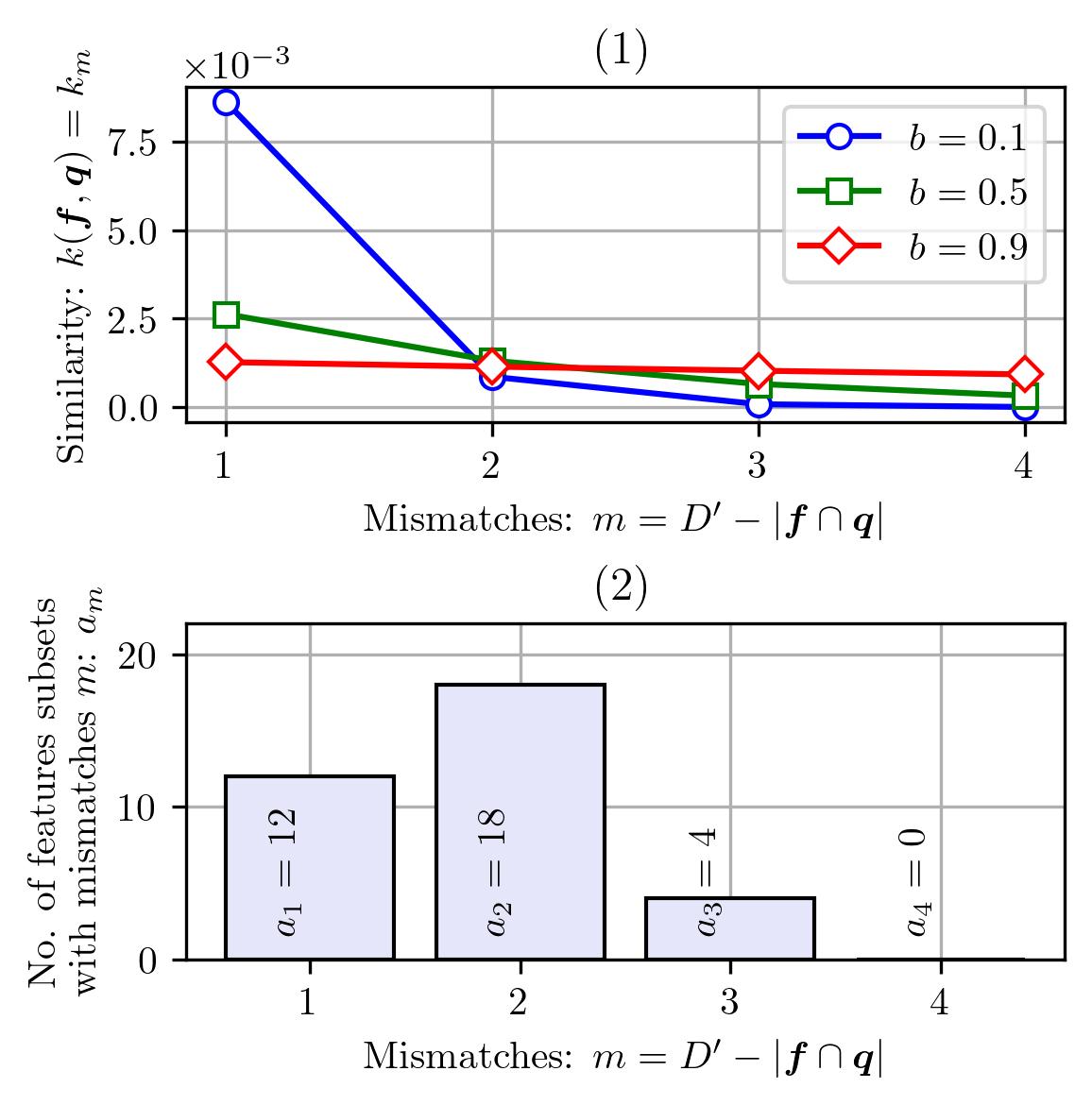}
 \caption{
 Relationship among mismatches $m$, similarity $k_m$, and the number of features subsets $a_m$ in FCS $\bm{\Omega}$. 
 Calculation conditions were $(D, D^\p, h) = (7, 4, 0.5)$.
 }
 \label{f_sim}
\end{figure}

A numerical example of the similarity function $k(\bm{f}, \bm{u})$ defined in Equation (\ref{eq_ma2}) is discusses in this section.
The relationship between $m$ and $k_m$ at $(D, D^\p, h) = (7, 4, 0.5)$ is illustrated in Fig. \ref{f_sim} (1).
We observed that the larger the number of mismatches $m$ between the feature subsets, the smaller the value of $k_m$.
Particularly, the smaller the value of $b$, the larger the decrease in similarity with respect to the number of mismatches.
Therefore, if matching multiple features are important, the value of $b$ should be smaller.
Otherwise, setting $b=0.5$ causes the similarity to decrease gradually with respect to an increase in the mismatch $m$.
Further, $a_m$ under the condition $(D, D^\p) = (7, 4)$ is shown in Fig. \ref{f_sim} (2).
No case showed four mismatches ($a_4 = 0$).
Assuming an overall feature set $\bm{F} = \{ f_1, \cdots, f_7 \}$ and a feature subset $\bm{u} = \{f_1, f_2, f_3, f_4 \}$ created from this set, a feature subset $\bm{f}$ with four mismatches can not be created.
Therefore, $a_4 = 0$ was satisfied.

\section*{Appendix 3: Calculation cost}
MAABO-MT uses parameters $N_\m{U}$ and $N_\m{E}$ to control the computation time (see Equation (\ref{eq_n_q_n_e})).
In this section, the effects of the parameters on reducing computation time are explained with reference to the experimental results.
As shown in Fig. \ref{f_nq}, by setting $N_\m{U}$ to approximately $D/5$, the search space size was approximately halved.
Therefore, we adopted $N_\m{U} = \lfloor D/5 \rfloor$.
Moreover, the sampling size $N_\m{E}$ has the effect of reducing computation time at the expense of search accuracy.
In other words, $N_\m{E}$ controlled the tradeoff between search accuracy and computational cost.
To investigate the relationship between the search performance and computation time, MAABO-MT was run with $N_\m{E} \in \{10^1, 10^2, 10^3, \infty\}$.
$N_\m{E} \leftarrow \infty$ indicated that all search targets were sampled and Bayesian optimization was performed.
Other parameters were the same as those used in Experiment 2.

\begin{figure*}[t]
 \centering
 \includegraphics[scale=0.82]{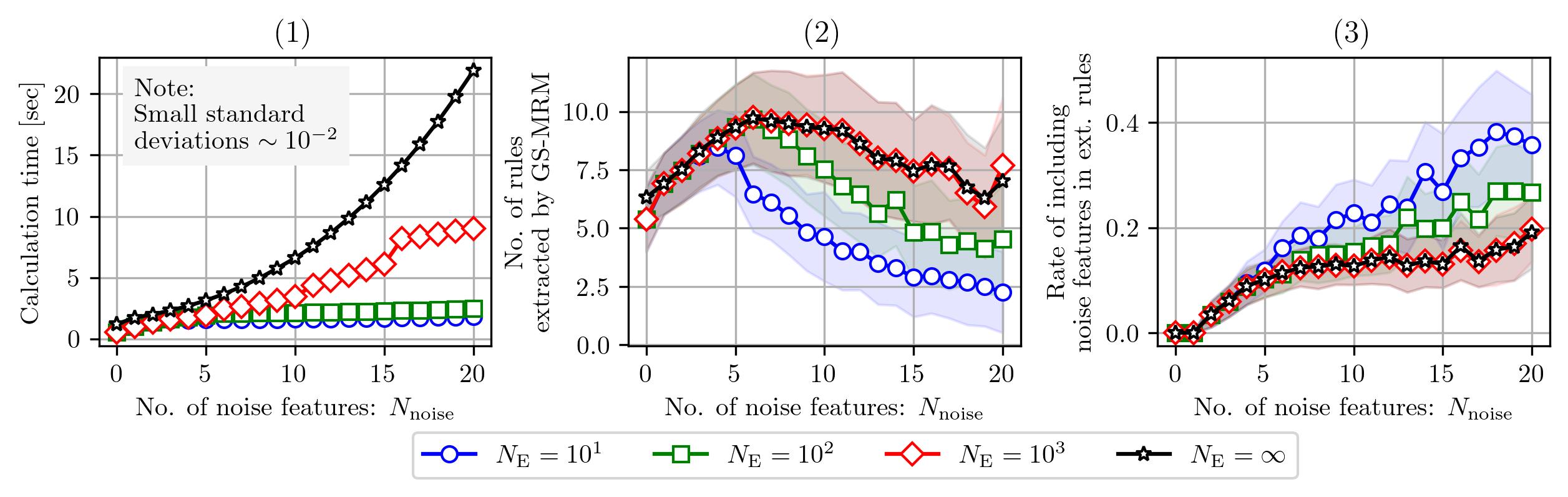}
 \caption{
 Relationship among sampling size $N_\m{E}$, computation time, and performance of extracting rules.
Values are the averages and the error bars are 0.5 std. of 50 random seeds.
 }
 \label{f_ex3}
\end{figure*}

The results obtained from the aforementioned procedure are shown in Fig. \ref{f_ex3}.
The horizontal axis represents the number of noise features added to the nine appropriate features of the Titanic dataset.
Fig. \ref{f_ex3} (1) presents the result of the computation time.
When $N_\m{E} \leftarrow \infty$, the computation time increased rapidly as the feature size increased because the size of the FCS $\bm{\Omega}$ was determined by a combination of $(D, D^\p)$ (see Equation (\ref{eq_comb_size})).
However, as $N_\m{E}$ decreased, the computation time was also reduced.
In particular, in the $N_\m{E} \in \{10^1, 10^2 \}$ cases, there was no significant increase in computation time even as the feature size increased.
Therefore, $N_\m{E}$ is a parameter that controls computation time.

Figs. \ref{f_ex3} (2) and (3) show the number of rules extracted by GS-MRM and the rate of noise features included in them, respectively.
By comparing $N_\m{E} \in \{10^1, 10^2 \}$ and $N_\m{E}= \infty$, the number of extracted rules decreases and the noise content increases by setting a small $N_\m{E}$.
Therefore, we can say that if $N_\m{E}$ is too small, the search performance of the rule mining decreases.
However, when comparing $N_\m{E} = 10^3$ and $N_\m{E} \leftarrow \infty$, the search performances were almost the same.
From Fig. \ref{f_ex3} (1), the computation time for $N_\m{E} = 10^3$ is approximately half that for $N_\m{E} \leftarrow \infty$.
Therefore, there exists $N_\m{E}$ such that only the computational cost can be reduced while maintaining the search performance.
This is an important result in situations in which rule mining must be performed using limited computational resources.

\end{document}